%% file: main.tex
\documentclass[10pt,twocolumn,letterpaper]{article}

\usepackage[pagenumbers]{wacv}
\usepackage[margin=1in]{geometry}

\input{preamble}

\definecolor{linkblue}{rgb}{0.21,0.49,0.74}
\usepackage[breaklinks,colorlinks,allcolors=linkblue]{hyperref}
\hypersetup{
  pdftitle={SCMerge: Spatial-Consistency-Aware Token Pair Merging for Efficient Semantic Segmentation},
  pdfauthor={Junfeng Ren; Yi Gong; Yuan Zeng}
}

\title{SCMerge: Spatial-Consistency-Aware Token Pair Merging for Efficient Semantic Segmentation}
\author{
  Junfeng Ren$^{1}$ \quad Yi Gong$^{1}$ \quad Yuan Zeng$^{2}$\\
  $^{1}$Southern University of Science and Technology\\
  $^{2}$Shenzhen Technology University\\
  {\tt\small \href{mailto:junfengren3253@gmail.com}{junfengren3253@gmail.com}}
}
\date{}

\begin{document}
\raggedbottom
\maketitle
\input{sec/0_abstract}
\input{sec/1_introduction}
\input{sec/2_related_work}
\input{sec/3_method}
\input{sec/4_experiments}
\input{sec/5_conclusion}

\FloatBarrier
{
  \small
  \bibliographystyle{ieeenat_fullname}
  \bibliography{main}
}
\end{document}

%% file: preamble.tex
\usepackage{microtype}
\usepackage{soul}
\usepackage{algorithm}
\usepackage{algorithmic}
\usepackage{multirow}
\usepackage{adjustbox}
\usepackage{placeins}

\renewcommand{\paragraph}[1]{\vspace{.2em}\noindent\textbf{#1.}}

\newcommand{\method}{SCMerge\xspace}

\setuldepth{foobar}

%% file: sec/0_abstract.tex
\begin{abstract}
Vision Transformer architectures have showcased impressive performance on semantic segmentation, yet their quadratic attention complexity hinders practical applications. Recent works have focused on merging tokens to reduce sequence length and thus improve transformer efficiency. However, existing token merging methods primarily rely on feature similarity to identify similar tokens, overlooking spatial consistency, which may introduce segmentation errors in semantic segmentation. In this work, we propose SCMerge, a training-free token merging method that jointly considers feature similarity and spatial consistency in pairing similar tokens. The pairwise tokens are then merged within the encoder, and the resulting token sequences are reconstructed onto the original patch grid after decoding. The proposed method is flexible and can be plugged into existing vision Transformer-based semantic segmentation pipelines. Extensive experiments on the benchmark datasets ADE20K, Pascal Context, and Cityscapes show that SCMerge consistently achieves a better trade-off between segmentation accuracy and efficiency than baseline methods.

\end{abstract}

%% file: sec/1_introduction.tex
\section{Introduction}
\label{sec:intro}
Semantic segmentation has made significant progress with the recent introduction of Vision Transformers (ViTs)~\cite{dosovitskiy2020image,strudel2021segmenter,zheng2021rethinking,ranftl2021vision}. However, ViTs with self-attention suffer from quadratic computational complexity and long latency, limiting their application to high-resolution images. A primary avenue for addressing this issue is to reduce token sequence length in self-attention and streamline computation~\cite{rao2021dynamicvit,liang2022not,bolya2022token,ryoo2021tokenlearner,yin2022avit,pan2021iared2,tang2022patch}. Although most existing token reduction methods work well for image classification, they often struggle in dense prediction tasks, such as semantic segmentation. The main challenge is to reduce the token sequence length while preserving the spatial and semantic details required by the segmentation decoder.

This work focuses on ViT-based semantic segmentation and aims to present a training-free token merging method to improve the efficiency of existing pipelines. Existing training-free token reduction methods have shown that merging similar tokens can effectively reduce the token sequence length and computational complexity. ToMe performs bipartite matching in feature space and merges similar tokens via averaging~\cite{bolya2022token}.
Mutual Pair Merging (MPM) employs mutual nearest-neighbor matching and combines the matched tokens, yielding computation and latency improvements without modifying segmentation decoders~\cite{rave2026mpm}.
In addition, ALGM employs a segmentation-specific local-to-global merging schedule that combines local and global token matching~\cite{norouzi2024algm}. Despite their potential, existing methods mainly match tokens based on feature similarity, which can cause segmentation errors when local feature content is similar but the tokens are spatially or globally different. Tokens near semantic boundaries may be incorrectly paired with interior tokens from neighboring regions, leading to boundary blurring. In addition, interior tokens from different objects can have similar features and be mistakenly identified as redundant. More recently, several works~\cite{lu2023content,huang2025tosa} have proposed to address this issue by introducing spatial information or learned policies, but at the cost of additional inputs or training.

In this paper, we propose SCMerge, a training-free token merging method to reduce the token length in ViT-based semantic segmentation pipelines, aiming to achieve a better trade-off between segmentation accuracy and computational cost. To guide the merging process, We introduce a token-pairing criterion that jointly considers local feature similarity and spatial consistency.
By merging only tokens that are similar in both feature and spatial consistency, this mechanism improves the efficiency–accuracy trade-off. Specifically, we first compute the cosine similarity between all token pairs. A local variance measure is then used to identify tokens that are near semantic boundaries or lie in a smooth interior region. For each candidate pair, we compute a spatial-consistency penalty that discourages merging of locally similar but boundary-sensitive tokens, as well as tokens belonging to different objects. Finally, paired tokens are merged via feature averaging, and the reduced token sequence is passed through the segmentation decoder before being mapped back to the original token grid. This design refines the token pairing process to preserve semantic boundaries and neighborhood spatial consistency, improving segmentation efficiency.

Our SCMerge does not require additional supervision, extra input modalities, or modifications to the segmentation head, allowing it to be plugged into ViT-based semantic segmentation pipelines. We evaluate the effectiveness of the proposed token merging module using ViT-based segmentation models~\cite{strudel2021segmenter} on three semantic segmentation datasets, including ADE20K~\cite{zhou2017scene}, Pascal Context~\cite{mottaghi2014role}, and Cityscapes~\cite{cordts2016cityscapes}. Experimental results demonstrate that plugging our SCMerge module into ViT-based semantic segmentation models improves the trade-off between computational complexity and segmentation accuracy. Our contributions can be summarized as follows:
\begin{itemize}
\item We propose a novel training-free token-pair merging method that incorporates a spatial-consistency penalty into feature-similarity-based matching. This method prevents the merging of locally similar but spatially incompatible tokens and improving the accuracy–efficiency trade-off.

\item We design a local feature variation metric to identify boundary-sensitive tokens and introduce a spatial-consistency penalty, comprising a boundary-sensitive term and a neighborhood consistency term, to reduce incorrect merging near semantic boundaries and preserve fine spatial detail.

\item Extensive experiments demonstrate that our method can be integrated into ViT-based semantic segmentation models to achieve a better accuracy‑efficiency trade‑off than baseline methods relying only on feature similarity.
\end{itemize}

%% file: sec/2_related_work.tex
\section{Related Work}
\label{sec:related-work}

\paragraph{Efficient ViT Segmentation}
Vision Transformer-based segmentation models, including SETR~\cite{zheng2021rethinking}, Segmenter~\cite{strudel2021segmenter}, DPT~\cite{ranftl2021vision}, and Mask2Former~\cite{cheng2022masked}, have achieved promising performance in dense prediction tasks, but they suffer from high computational costs when processing high-resolution images. A possible solution is to redesign the architecture or attention mechanism.
SegFormer~\cite{xie2021segformer} adopts a hierarchical Transformer encoder with a lightweight all-MLP decoder.
Swin Transformer~\cite{liu2021swin} restricts self-attention to shifted local windows, and PVT~\cite{wang2021pyramid} progressively reduces spatial resolution to build pyramid features.
FlashAttention~\cite{dao2022flashattention} accelerates attention by reducing memory input and output between GPU high-bandwidth memory and on-chip memory through tiled computation.
This work focuses on reducing redundant tokens inside a pretrained ViT encoder
, preserving the compact representation through decoding, and reconstructing the dense prediction grid
without changing the segmentation head.

\paragraph{Token Reduction in Vision Transformers}
Many recent methods accelerate ViTs by reducing the token sequences passed to subsequent layers.
DynamicViT~\cite{rao2021dynamicvit} learns to prune uninformative tokens dynamically, while TokenLearner~\cite{ryoo2021tokenlearner} learns a compact set of informative tokens.
A-ViT~\cite{yin2022avit} introduces adaptive token halting, and IA-RED$^2$~\cite{pan2021iared2} reduces redundancy with interpretability-aware token selection.
Patch Slimming~\cite{tang2022patch} and SPViT~\cite{kong2022spvit} study efficient token pruning, while ATS~\cite{fayyaz2022adaptive} samples tokens adaptively.
Token Pooling~\cite{marin2023token_pooling}, STViT~\cite{huang2023stvit}, HeatViT~\cite{dong2023heatvit}, and DiffRate~\cite{chen2023diffrate} further explore pooling, sampling, pruning, or learnable compression rates.
These methods %
either target image classification or require learned policies, retraining, or architecture-specific modifications. Unlike classification tasks, preserving thin structures and fine boundaries is critical in dense prediction tasks. This depends on having a sufficient number of tokens for accurate reconstruction and precise predictions.
This work introduces a training-free token merging method designed for dense prediction.

\paragraph{Training-free Token Merging}
Token merging has emerged as an effective technique for improving the efficiency of ViTs. By combining similar tokens rather than discarding them, token merging methods substantially reduce computational complexity and latency in image classification and semantic segmentation.
ToMe~\cite{bolya2022token} is a training-free token merging method that matches feature-similar tokens and averages them progressively inside the encoder.
Several works have extended token merging with learned or adaptive mechanisms, including learned patch merging~\cite{renggli2022learning}, efficient token merging~\cite{feng2023efficient}, delayed spatial merging~\cite{heo2023delayed}, and learned pruning-and-merging thresholds~\cite{bonnaerens2023learned}.
In addition, fusion-based or saliency-driven methods, such as token fusion~\cite{kim2024tofu}, multi-criteria fusion~\cite{lee2024multi}, spectrum-preserving merging~\cite{tran2024accelerating}, learnable token merging~\cite{wang2024learnable}, and saliency-driven merging~\cite{xie2026saliency} have been studied for token merging.
Mutual Pair Merging (MPM)~\cite{rave2026mpm} forms mutual nearest-neighbor token pairs and preserves a reconstruction map before the dense prediction decoder.
These methods can significantly improve ViT efficiency by averaging similar tokens, but their token grouping strategies mainly rely on feature similarity, which neglects spatial consistency when measuring token similarity. %
Our \method addresses this limitation by using local feature variation to pair tokens in a more spatial-aware manner.

\begin{figure*}[!t]
\centering
\includegraphics[width=.95\textwidth]{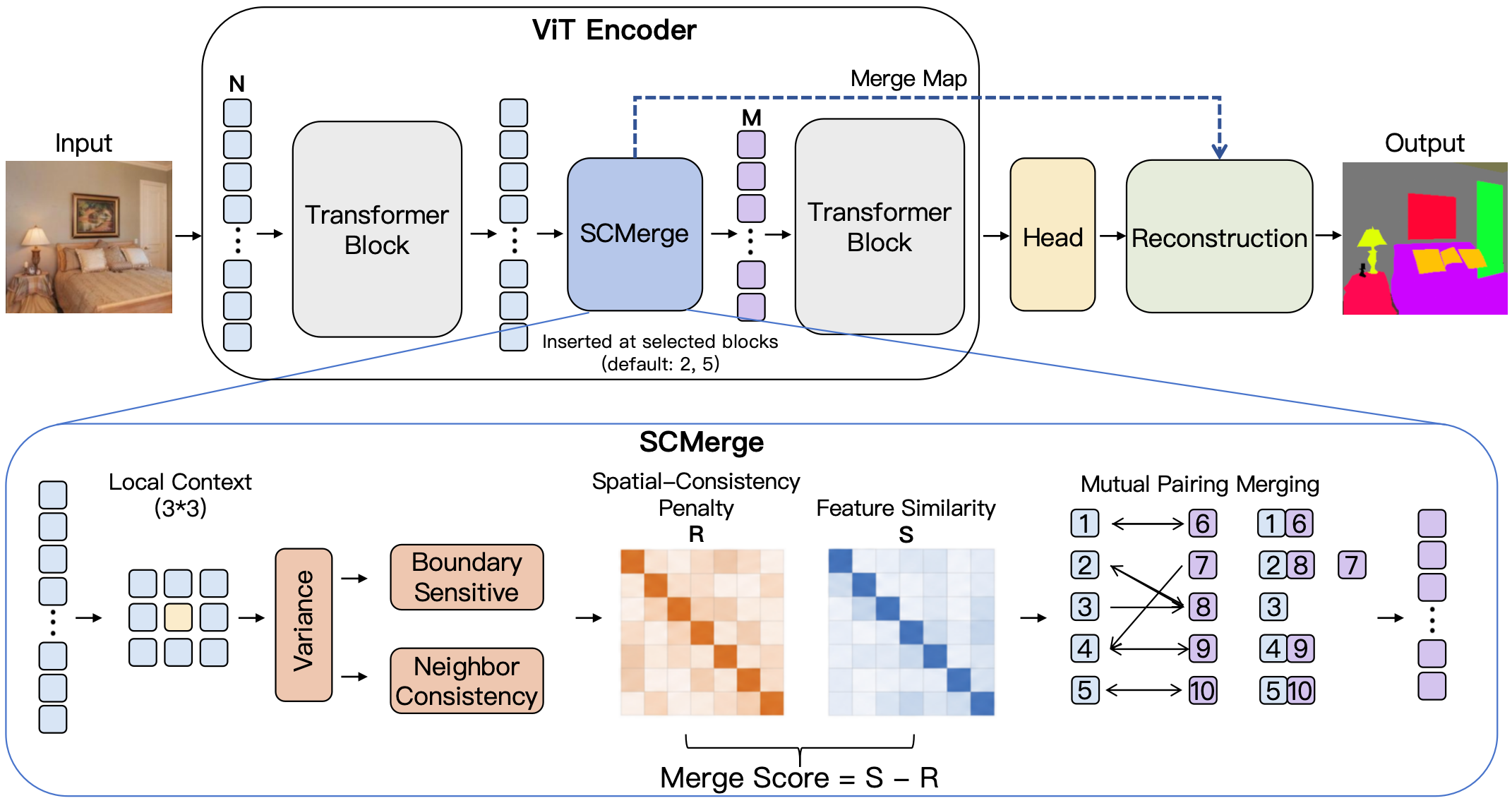}
\caption{Overview of SCMerge. SCMerge is a plug-and-play module that can be inserted into multiple ViT encoder blocks to reduce the token sequence length while maintaining a merge map. It combines feature similarity with a spatial-consistency penalty to guide similar token pairing and merging. The reduced token sequence remains compact throughout decoding and is reconstructed onto the original dense prediction grid only after the decoder.}
\label{fig:framework}
\end{figure*}

\paragraph{Spatial-aware Token Merging for Semantic Segmentation}
Several recent methods have incorporated task-specific or spatial information to improve token merging for semantic segmentation.
CTS~\cite{lu2023content} learns a content-aware token-sharing policy to preserve semantic structures.
ALGM~\cite{norouzi2024algm} designs an adaptive local-to-global merging schedule for segmentation backbones.
ToSA~\cite{huang2025tosa} introduces spatial awareness into token merging so that merging decisions consider both visual feature similarity and spatial information.
These methods demonstrate the importance of spatial information for semantic segmentation, but they often rely on learned policies, specialized schedules, additional spatial information, or method-specific implementations.
\method follows the same motivation of spatial-aware token merging, but it is training-free and can be plugged into existing ViT-based semantic segmentation pipelines.

%% file: sec/3_method.tex
\section{Method}
\label{sec:method}

As illustrated in Figure~\ref{fig:framework}, SCMerge is a training-free module that can be inserted into multiple blocks of a ViT encoder to reduce the token sequence length from $N$ to $M$ ($M< N$),
while maintaining a merge map for reconstructing compact segmentation head outputs onto the original token grid after decoding.
Let $X=[x_1,\ldots,x_N]\in\mathbb{R}^{N\times d}$ denote the image tokens at a merge layer, where $N$ is the number of input tokens and $d$ is the feature dimension.
Each token $x_i$ is indexed by a representative coordinate $c_i$ for local neighborhood statistics, while a merge map tracks reconstruction assignments across merging stages. At each insertion, SCMerge first computes a feature-similarity matrix $S$ to measure content similarity between tokens. It then calculates local context variation within a $3\times3$ token neighborhood and uses this variation to assign a spatial-consistency penalty score when forming matching pairs. The matched token pairs are merged into a reduced sequence $Y=[y_1,\ldots,y_M]\in\mathbb{R}^{M\times d}$. The merged tokens are then propagated through the remaining backbone layers, while the corresponding merge map is stored.
The merged tokens are then fed into the decoder and segmentation head of the original ViT-based model to produce compact prediction logits. Finally, the merge map is used to reconstruct these compact logits onto the final semantic segmentation results.

\subsection{Preliminaries}
We follow existing semantic segmentation models~\cite{strudel2021segmenter} and use a standard Vision Transformer (ViT) encoder~\cite{dosovitskiy2020image} and Mask Transformer decoder as our backbone. Given an input image of size $H\times W$ and patch size $P$, the image is divided into $N_0=H/P\times W/P$ non-overlapping patches. Each patch is linearly projected into a $d$-dimensional embedding, which is added to a learned positional embedding, resulting in the initial token sequence $X_0\in\mathbb{R}^{N_0\times d}$. These tokens are then processed by a stack of Transformer blocks consisting of self-attention and feed-forward layers.

The decoder then jointly processes the tokens from the encoder together with $C$ learnable class embeddings, where $C$ denotes the number of semantic classes.
After the Transformer decoder layers, the output image and class embeddings are projected and normalized, and the pairwise inner products between the image embeddings and the class embeddings produce patch-level class logits.
In the original full-token model, these logits correspond to the $N_0$ patch locations and are reshaped into an $H/P\times W/P$ patch grid. They are then upsampled to the input resolution via bilinear interpolation.

\textbf{SCMerge Insertion.}
SCMerge is inserted after selected ViT encoder blocks and operates on the current token sequence. At each insertion, SCMerge reduces the token sequence length from $N$ to $M$($M<N$).
To support multiple merge layers, SCMerge maintains a representative coordinate for each token and a merge map. The coordinate indicates the token's position in the sparse token layout used for variation computation, while the merge map records, for each original patch location, the token that represents it. The merge map is updated after each merge stage. The reduced token sequence is then propagated through the remaining encoder blocks and passed through the Mask Transformer decoder. The decoder architecture and parameters remain unchanged, while only the number of input tokens is reduced. After decoding, the merge map restores the reduced logits to the original patch grid, followed by bilinear upsampling to the input resolution.

\subsection{Spatial-Consistency-Aware Token Merging}
\textbf{Local Feature Variation.} Following existing training-free token merging methods~\cite{bolya2022token,rave2026mpm}, we use cosine similarity to measure token similarity.
For each normalized token feature
$\bar{x}_i=x_i/(\|x_i\|_2+\epsilon)$,
the pairwise token similarity is given by
\begin{equation}
S_{ij}=\bar{x}_i^\top\bar{x}_j,
\quad i,j\in\{1,\ldots,N\},
\label{eq:feature_similarity}
\end{equation}
$S_{ij}$ denotes the pairwise feature similarity between tokens $i$ and $j$, and the diagonal entries of $S$ are set to $-\infty$ to avoid self-matching.
To estimate local variation efficiently, we form a deterministic $d_s$-channel sketch by averaging contiguous groups of the normalized feature channels.
The sketches are placed on the current sparse token layout, where the channel-wise variance within each $3\times3$ neighborhood is averaged to obtain a scalar variation $v_i$.
The sketch is used only for local-variation estimation, while token similarity and merging retain the full-dimensional features.
We then apply min--max normalization across the active tokens:
\begin{equation}
e_i=\frac{v_i-\min_k v_k}{\max_k v_k-\min_k v_k+\epsilon}.
\label{eq}
\end{equation}
$e_i\in[0,1]$ represents a normalized local variation score that quantifies the feature variation around $i$.
A low $e_i$ indicates a spatially homogeneous neighborhood, while a high $e_i$ indicates strong local feature variation, typically occurring near semantic boundaries, thin structures, or heterogeneous regions.

\textbf{Spatial-Consistency-Aware Token Pairing.}
We first define a score to detect token pairs that may cross semantic boundaries, that is,
\begin{equation}
B_{ij} = \max(e_i,e_j).
\label{eq:boundary_risk}
\end{equation}
Since $B_{ij}$ takes the maximum local variation of the two endpoints, it grows when either token lies in a high-variation region, such as near boundaries or thin structures. This penalty term discourages boundary tokens from being merged with interior tokens, thus reducing boundary blurring and preserving spatial detail.

Next, we define a neighborhood consistency score between candidate tokens $i$ and $j$ as the absolute difference of their local variation values:
\begin{equation}
D_{ij} = |e_i-e_j|,
\label{eq:neighbor_risk}
\end{equation}
where $D_{ij}$ measures the difference in local feature variation between tokens $i$ and $j$.
Even if two tokens have similar local content, a large $D_{ij}$ implies they belong to different structural regions and should not be merged. The final spatial-consistency penalty is defined as the Euclidean norm of the boundary-sensitivity term $B_{ij}$ and the neighborhood consistency term $D_{ij}$:

\begin{equation}
R_{ij} =
\sqrt{B_{ij}^{2}+D_{ij}^{2}}.
\label{eq:context_risk}
\end{equation}

The merge score is defined by subtracting the spatial-consistency penalty from the feature similarity:
\begin{equation}
A_{ij} = S_{ij} - R_{ij},
\label{eq:context_affinity}
\end{equation}
where $A_{ij}$ is the final merge score used for mutual pair selection.
A high $A_{ij}$ indicates that tokens $i$ and $j$ are feature-similar and that their merge is spatially consistent (i.e., the pair is less likely to involve boundary-sensitive or locally inconsistent regions). The merge score matrix $A$ is used to determine which pairs of tokens are similar and can be merged. For each token $i$, we find its most similar neighbor token $j$ by selecting the one with the highest merge score:
\begin{equation}
n(i)=\arg\max_{j\neq i} A_{ij}.
\label{eq:nearest_neighbor}
\end{equation}
The most similar token is selected as the maximum $A_{ij}$. Since the pair selection only relies on the ranking of $A_{ij}$, $A_{ij}$ is not constrained to be non-negative. As with the similarity matrix $\mathbf{S}$, the diagonal of $\mathbf{A}$ is set to $-\infty$ to avoid self-matching. As in~\cite{rave2026mpm}, we define that a pair $(i,j)$ is selected only when the nearest-neighbor relation is reciprocal:
\begin{equation}
(i,j)\in\mathcal{P}
\Longleftrightarrow
n(i)=j \ \mathrm{and}\ n(j)=i,
\label{eq:mutual_pair}
\end{equation}
where $\mathcal{P}$ denotes the set of selected mutual pairs. Tokens that do not belong to any selected pair remain as singletons.

\textbf{Token Pair Merging.} For each selected token pair ($i$, $j$), the two tokens are merged by averaging their features, producing a new token:
\begin{equation}
y=\frac{x_i+x_j}{2}.
\label{eq:supp_uniform_merge}
\end{equation}
The merged token retains its representative coordinate, keeping all active coordinates on the original patch grid. Unpaired tokens are kept unchanged as singletons. The resulting token sequence is $Y=[y_1,\ldots,y_M]\in\mathbb{R}^{M\times d}$, where $M$ denotes the total number of tokens after merging. Algorithm~\ref{alg:scmerge} summarizes the overall procedure of SCMerge.

\begin{algorithm}[t]
\caption{SCMerge at one insertion}
\label{alg:scmerge}
\small
\textbf{Input:} token sequence $X=[x_1,\ldots,x_N]$, representative coordinates, merge map\\
\textbf{Output:} reduced sequence $Y=[y_1,\ldots,y_M]$, representative coordinates, merge map
\begin{algorithmic}[1]
\STATE Compute the feature similarity $S_{ij}=\bar{x}_i^{\top}\bar{x}_j$, where $\bar{x}_i=x_i/(\|x_i\|_2+\epsilon)$
\STATE Compute $v_i$ from the $d_s$-channel sketch over the $3\times3$ neighborhood and normalize $e_i=\frac{v_i-\min_k v_k}{\max_k v_k-\min_k v_k+\epsilon}$
\STATE Compute the spatial-consistency penalty $R_{ij}=\sqrt{\max(e_i,e_j)^{2}+|e_i-e_j|^{2}}$ and the merge score $A_{ij}=S_{ij}-R_{ij}$
\STATE Find $n(i)=\arg\max_{j\neq i}A_{ij}$ and select mutual pairs
$\mathcal{P}=\{(i,j)\mid n(i)=j,\ n(j)=i\}$
\STATE Merge each pair by $y=(x_i+x_j)/2$ and retain its representative coordinate; keep unmatched tokens
\STATE return $Y$, representative coordinates, and merge map
\end{algorithmic}
\end{algorithm}

\subsection{Decoding and Reconstruction}
\label{sec:Decoder}
The reduced token sequence $Y$ is passed through the segmentation decoder without spatial expansion, and the original token grid is reconstructed only after decoding for dense prediction. The segmentation decoder processes the merged token sequence $Y$ and produces class logits $H\in\mathbb{R}^{M\times C}$, where $C$ is the number of semantic classes. The merge map maps each merged token’s logits back to the original patch locations, reconstructing the $N_0$ patch logits. These logits are then reshaped into the original patch grid and bilinearly upsampled to the input resolution. For the Mask Transformer decoder, the $M$ reduced tokens are jointly processed with $C$ class embeddings, resulting in a sequence of $M+C$ tokens.
Thus, the token reduction is preserved across the entire encoder–decoder pipeline.
The decoder remains unchanged in architecture and parameters, introducing no extra learnable parameters.

%% file: sec/4_experiments.tex
\section{Experiments}
\label{sec:experiments}
This section evaluates the effectiveness and generalization of our method on semantic segmentation. We plug our token merging method into existing ViT-based semantic segmentation models and present quantitative and qualitative comparisons on three benchmark semantic segmentation datasets among full-token baseline models, models with feature-similarity-based token merging, and models with our method. Later, we also provide a comprehensive ablation study supporting our design choices. %
\begin{table*}[!t]
\centering
\captionsetup{skip=3pt}
\caption{Quantitative comparison results on ADE20K, Pascal Context, and Cityscapes with Segmenter backbones.}
\label{tab}
\small
\renewcommand{\arraystretch}{0.96}
\setlength{\tabcolsep}{4.5pt}
\begin{tabular*}{.98\textwidth}{@{\extracolsep{\fill}}lccccc@{}}
\toprule
Model/Method & mIoU & $N_{\mathrm{final}}$ & GFLOPs & Latency (ms) & $R$ $\uparrow$ \\
\midrule
\multicolumn{6}{l}{\textbf{ADE20K}} \\
\textbf{Segmenter-B/16}
& 48.48
& 1024.00
& 258.80
& 29.77
& -- \\
ToMe ($r{=}200$)
& 48.01
& 624.00
& 193.65
& 20.13
& 20.51 \\
MPM
& 47.81
& 563.29
& 182.75
& \textbf{19.58}
& 15.21 \\
MPM-top (76\%)
& 48.22
& 655.46
& 198.36
& 22.90
& 26.42 \\
\method
& \textbf{48.42}
& 653.98
& \textbf{181.21}
& 20.82
& \textbf{149.17} \\
\midrule
\textbf{Segmenter-B/8}
& 49.57
& 4096.00
& 1319.38
& 158.80
& -- \\
ToMe
& 49.15
& 2496.00
& 854.96
& 100.33
& 139.21 \\
MPM
& 49.20
& 2247.40
& \textbf{802.61}
& \textbf{91.81}
& 181.05 \\
MPM-top
& 49.26
& 2621.96
& 911.04
& 110.23
& 156.68 \\
\method
& \textbf{49.38}
& 2639.09
& 908.26
& 107.86
& \textbf{268.11} \\
\midrule
\textbf{Segmenter-L/16}
& 51.81
& 1600.00
& 1343.32
& 134.13
& -- \\
ToMe
& 51.48
& 976.00
& 886.72
& 80.58
& 162.27 \\
MPM
& 51.02
& 876.40
& \textbf{818.72}
& \textbf{73.89}
& 76.25 \\
MPM-top
& 51.45
& 1020.21
& 919.94
& 89.29
& 124.56 \\
\method
& \textbf{51.64}
& 1040.61
& 892.82
& 87.43
& \textbf{274.71} \\
\specialrule{1.2pt}{1.5pt}{1.5pt}
\multicolumn{6}{l}{\textbf{Pascal Context}} \\
\textbf{Segmenter-L/16}
& 58.08
& 900.00
& 685.67
& 56.89
& -- \\
ToMe
& 57.74
& 548.00
& 461.34
& 38.69
& 53.53 \\
MPM
& 57.46
& 486.28
& \textbf{418.49}
& \textbf{34.87}
& 35.52 \\
MPM-top
& 57.78
& 567.75
& 470.45
& 41.00
& 52.97 \\
\method
& \textbf{57.94}
& 581.48
& 456.49
& 39.51
& \textbf{124.14} \\
\specialrule{1.2pt}{1.5pt}{1.5pt}
\multicolumn{6}{l}{\textbf{Cityscapes}} \\
\textbf{Segmenter-S/16}
& 76.35
& 2304.00
& 232.07
& 35.98
& -- \\
ToMe
& 75.43
& 1404.00
& 161.97
& 24.90
& 12.04 \\
MPM
& 75.30
& 1334.67
& \textbf{159.70}
& \textbf{24.42}
& 11.01 \\
MPM-top
& 75.52
& 1530.32
& 174.66
& 28.72
& 8.75 \\
\method
& \textbf{75.68}
& 1565.60
& 162.22
& 26.52
& \textbf{14.12} \\
\midrule
\textbf{Segmenter-L/16}
& 79.01
& 2304.00
& 2010.69
& 208.92
& -- \\
ToMe
& 78.03
& 1404.00
& 1266.86
& 125.36
& 85.27 \\
MPM
& 77.72
& 1299.06
& \textbf{1190.04}
& \textbf{118.68}
& 69.95 \\
MPM-top
& 78.23
& 1498.76
& 1346.05
& 133.54
& 96.64 \\
\method
& \textbf{78.48}
& 1535.59
& 1337.82
& 132.80
& \textbf{143.62} \\
\bottomrule
\end{tabular*}
\end{table*}

\begin{figure*}[!t]
\centering
\includegraphics[width=.88\textwidth]{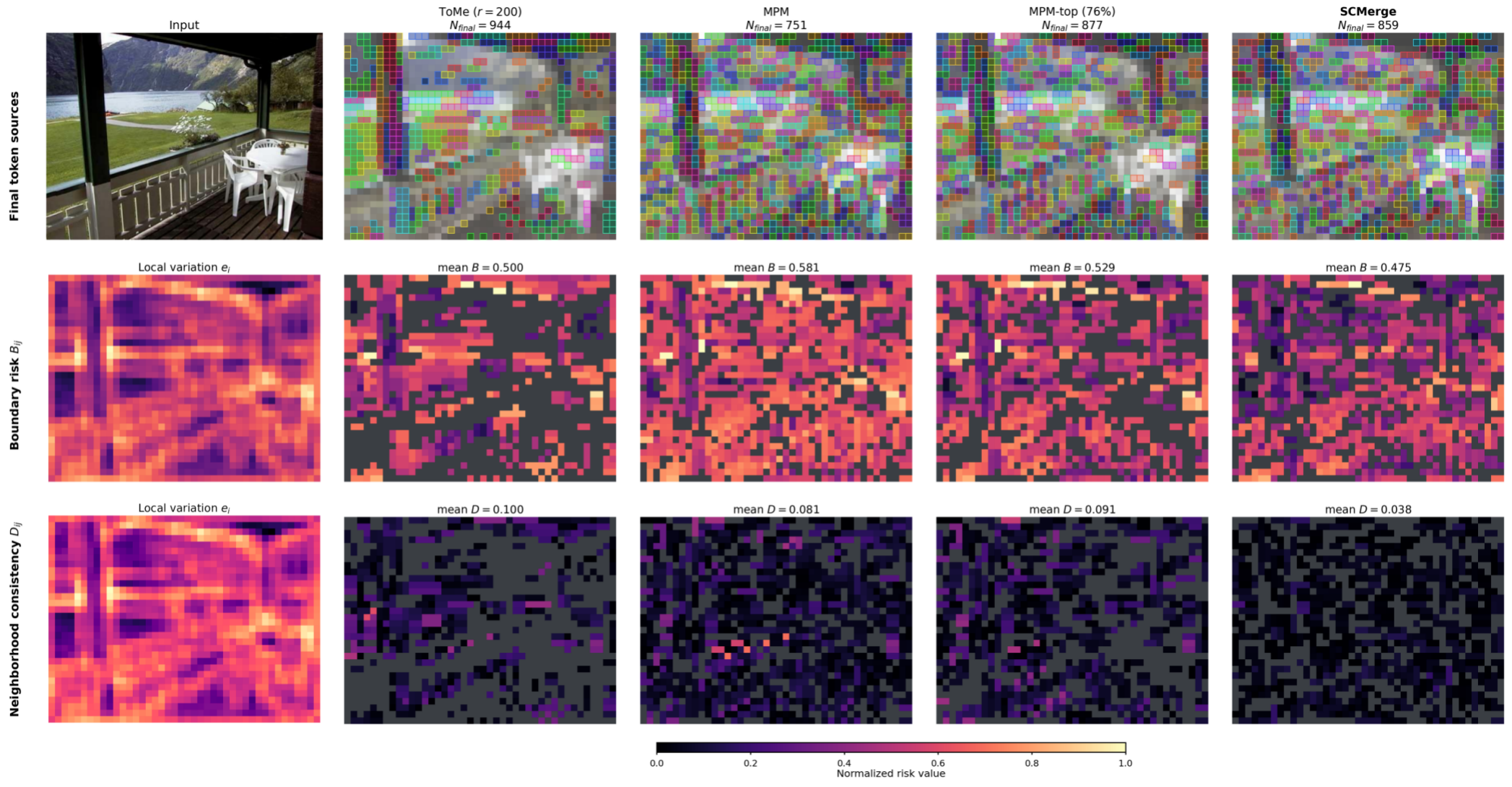}
\caption{Token-reduction and spatial-consistency visualization.
\method selects spatial-consistency token pairs with lower boundary sensitivity $B_{ij}$ and neighborhood consistency $D_{ij}$ than competing methods.}
\label{fig:qual_merge_main}
\end{figure*}

\begin{figure*}[!t]
\centering
\includegraphics[width=.88\textwidth]{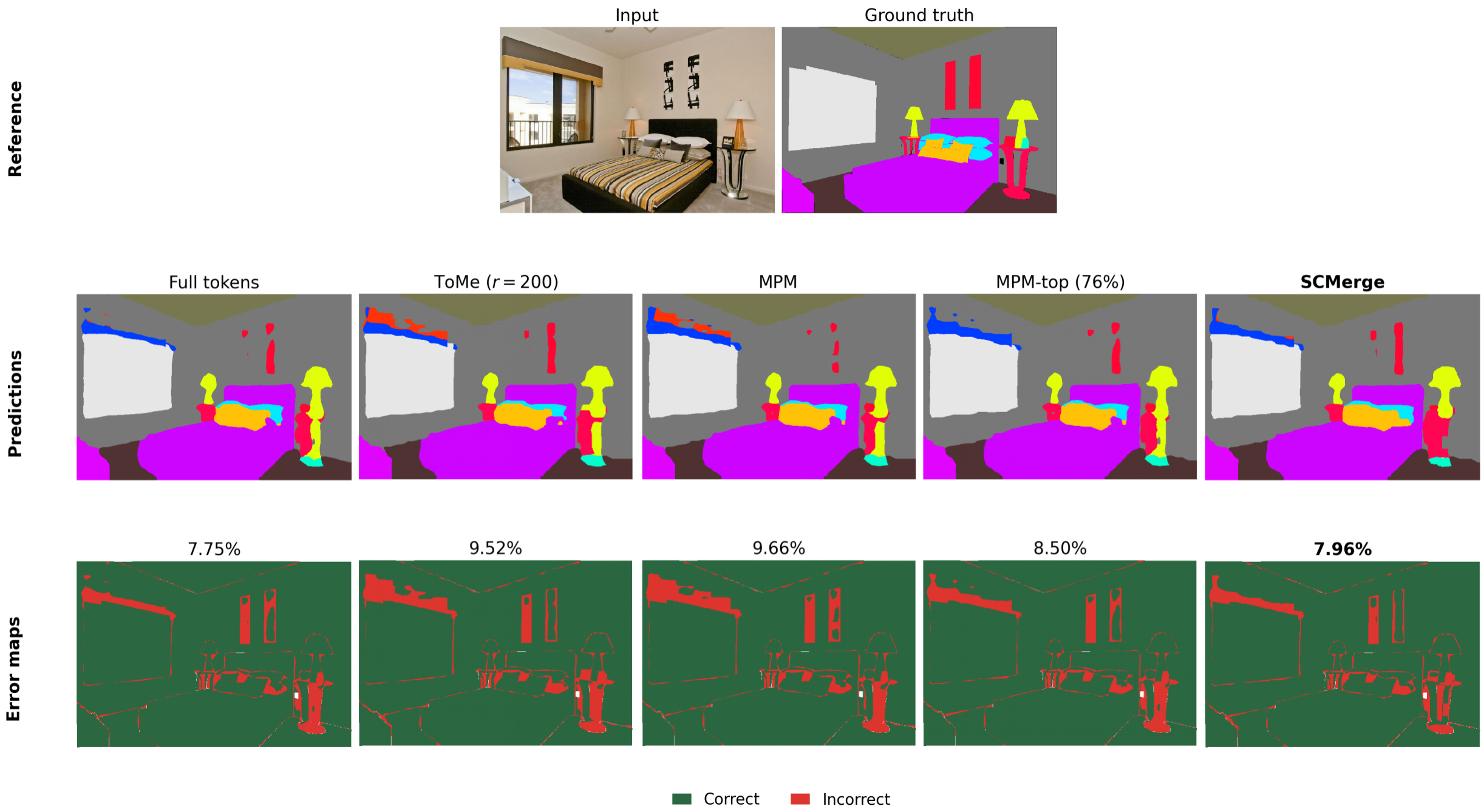}
\caption{Qualitative segmentation comparison on ADE20K.
Prediction and error maps show that \method produces fewer segmentation errors than the baseline token-merging methods, achieving the lowest pixel error rate.}
\label{fig:qual_error}
\end{figure*}

\subsection{Experimental Setup}

\textbf{Datasets and Models.}
We conduct experiments on three datasets: ADE20K~\cite{zhou2017scene}, Pascal Context~\cite{mottaghi2014role}, and Cityscapes~\cite{cordts2016cityscapes}, which respectively cover scene parsing, object and stuff segmentation, and high-resolution street scenes.
Our experiments use Segmenter~\cite{strudel2021segmenter} as a baseline full-token model. To provide a fair comparison, we retain its released encoder and Mask Transformer decoder and keep the input resolution and validation protocol of each checkpoint unchanged. We insert the token merging operation after the second and fifth encoder blocks by default.

\textbf{Implementation Details.}
We fix the deterministic channel-sketch width to $d_s=48$ across all datasets and backbones, using zero padding only when the feature dimension is not divisible by $d_s$.
After each merge stage, the remaining tokens are scattered onto a zero-filled sparse grid at their retained representative coordinates; only token features are averaged during merging.
We use zero-padded $3\times3$ average pooling with a fixed denominator to compute the first and second moments, and derive the variance as the centered second moment.
Assignment maps are combined across merge stages and used to reconstruct the dense logits after decoding the reduced tokens.

\textbf{Baselines.}
We use the officially released Segmenter checkpoint as the full-token baseline model and as the backbone of all training-free token merging methods: ToMe~\cite{bolya2022token}, MPM~\cite{rave2026mpm}, and \method. For ToMe~\cite{bolya2022token}, we use \(r=200\) for the 1024-token Segmenter-B/16 setting and scale \(r\) proportionally with the initial token count for other resolutions and backbones, preserving the same per-layer reduction ratio. To compare our method with MPM under a similar token budget, we configure MPM to merge the top 76\% of token pairs, and refer to this configuration as MPM-top.

\textbf{Evaluation Metrics.}
We report mIoU, the average final token count $N_{\mathrm{final}}$,
GFLOPs, and latency. Since token reduction is input-dependent,
$N_{\mathrm{final}}$ and GFLOPs are averaged over the entire validation
set. Latency is measured on an NVIDIA RTX 3090 with batch size 1 in
FP32 with TF32 disabled. We time the complete model forward pass,
including merging and reconstruction but excluding data loading and
preprocessing. After 50 warm-up iterations, the full validation set is
evaluated five times with explicit CUDA synchronization, and the mean
per-image latency is reported. All methods use the same hardware and
software environment.
Moreover, to quantify the trade-off between latency improvement and
accuracy drop, we define
\begin{equation}
R=
\frac{\mathrm{Latency}_{\mathrm{full}}-\mathrm{Latency}_{\mathrm{method}}}
{\mathrm{mIoU}_{\mathrm{full}}-\mathrm{mIoU}_{\mathrm{method}}},
\end{equation}
where the numerator denotes the latency reduction relative to the
full-token baseline and the denominator denotes the corresponding mIoU
decrease. A larger $R$ indicates greater latency reduction per unit of
accuracy loss. We report $R$ jointly with the corresponding raw mIoU
and latency values.

\subsection{Quantitative Comparison}

Table~\ref{tab} reports quantitative comparison results between the full-token baseline and token-merging methods in terms of mIoU, $N_{\mathrm{final}}$, GFLOPs, latency and accuracy-latency trade-off ratio $R$. We observe that all training-free methods can be plugged into existing segmentation models and improve their accuracy-latency trade-off. MPM achieves the lowest latency but sacrifices more segmentation accuracy than other token merging methods. Under a matched token budget, MPM-top achieves better segmentation accuracy than the original MPM but leads to higher latency.  Our \method, with spatial-consistency-aware token merging, achieves the best accuracy-latency trade-off among all training-free token merging methods. Specifically, on ADE20K, our \method consistently achieves the highest mIoU and the highest $R$ among the training-free methods, showing that its accuracy--latency advantage is maintained as token density and backbone capacity increase. The same trend transfers to Pascal Context and Cityscapes. Although MPM generally achieves lower raw latency, \method consistently preserves higher segmentation accuracy and yields a higher $R$, indicating a more favorable accuracy--latency trade-off. Across all six model--dataset settings, \method improves over MPM by $0.18$--$0.76$ mIoU while reducing full-token latency by approximately $26\%$--$36\%$, without retraining or dataset-specific fine-tuning.

\subsection{Qualitative Comparison}

In Figure~\ref{fig:qual_merge_main}, we visualize the token assignment maps that indicate the source patches for each merged token, along with the spatial-consistency penalty values of the selected pairs. Our \method achieves the lowest average values of both $B_{ij}$ and $D_{ij}$ across all compared methods. This result is consistent with the intended effect of the spatial-consistency penalty, which prioritizes pairs with low boundary sensitivity and small neighborhood inconsistency. The corresponding visualization shows that the proposed criterion encourages more spatially consistent token merging.

Figure~\ref{fig:qual_error} provides a visual comparison of the resulting segmentation predictions and the corresponding pixel-wise error maps. The visualization results show that our \method achieves higher segmentation accuracy  and a fewer pixel-wise error rate that baseline token-merging methods: ToMe, MPM, and MPM-top, with an error rate of $7.96\%$ compared to $9.52\%$, $9.66\%$, and $8.50\%$, respectively. This is consistent with the quantitative comparison results, demonstrating that spatial-consistency-aware token pairing is effective in preserving segmentation quality after token reduction.

\subsection{Ablation}
Table~\ref{tab:component_ablation_main} provides the effectiveness of spatial-consistency-aware token pairing and post-decoder reconstruction. We compare our full design with three variants: 1) without both spatial-consistency penalty $R_{ij}$ and post-decoder reconstruction, i.e., reconstructing the original token grid before the decoder (\emph{w/o penalty and Post-decoder}); 2) without spatial-consistency penalty $R_{ij}$ (\emph{w/o penalty}); and 3) without post-decoder reconstruction (\emph{w/o Post-decoder}). We observe that the spatial-consistency penalty improves segmentation accuracy while slightly increasing latency, whereas post-decoder reconstruction significantly reduces latency but slightly reduces accuracy. Our full design with both spatial-consistency penalty and post-decoder reconstruction achieves a better overall accuracy-latency trade-off. %

\begin{table}[!t]
\centering
\captionsetup{skip=4pt}
\caption{Quantitative ablation study of spatial-consistency penalty and post-decoder reconstruction on ADE20K with Segmenter-B/16
}
\label{tab:component_ablation_main}
\footnotesize
\setlength{\tabcolsep}{3.2pt}
\begin{tabular*}{\columnwidth}{@{\extracolsep{\fill}}lccc@{}}
\toprule
Variant & mIoU & Latency & $R$\\
\midrule
Full tokens & 48.48 & 29.77 & --\\
\midrule
\shortstack[l]{w/o penalty and\\Post-decoder Reconstruction} & 47.69 & 19.97 & 12.41 \\
w/o penalty & 47.56  & \textbf{16.63} & 14.28\\
w/o Post-decoder & \textbf{48.44} & 24.01  & 144.00\\
Full model & 48.42 & 20.82 & \textbf{149.17}\\
\bottomrule
\end{tabular*}
\end{table}

\begin{figure}[!t]
\centering
\includegraphics[width=.98\linewidth]{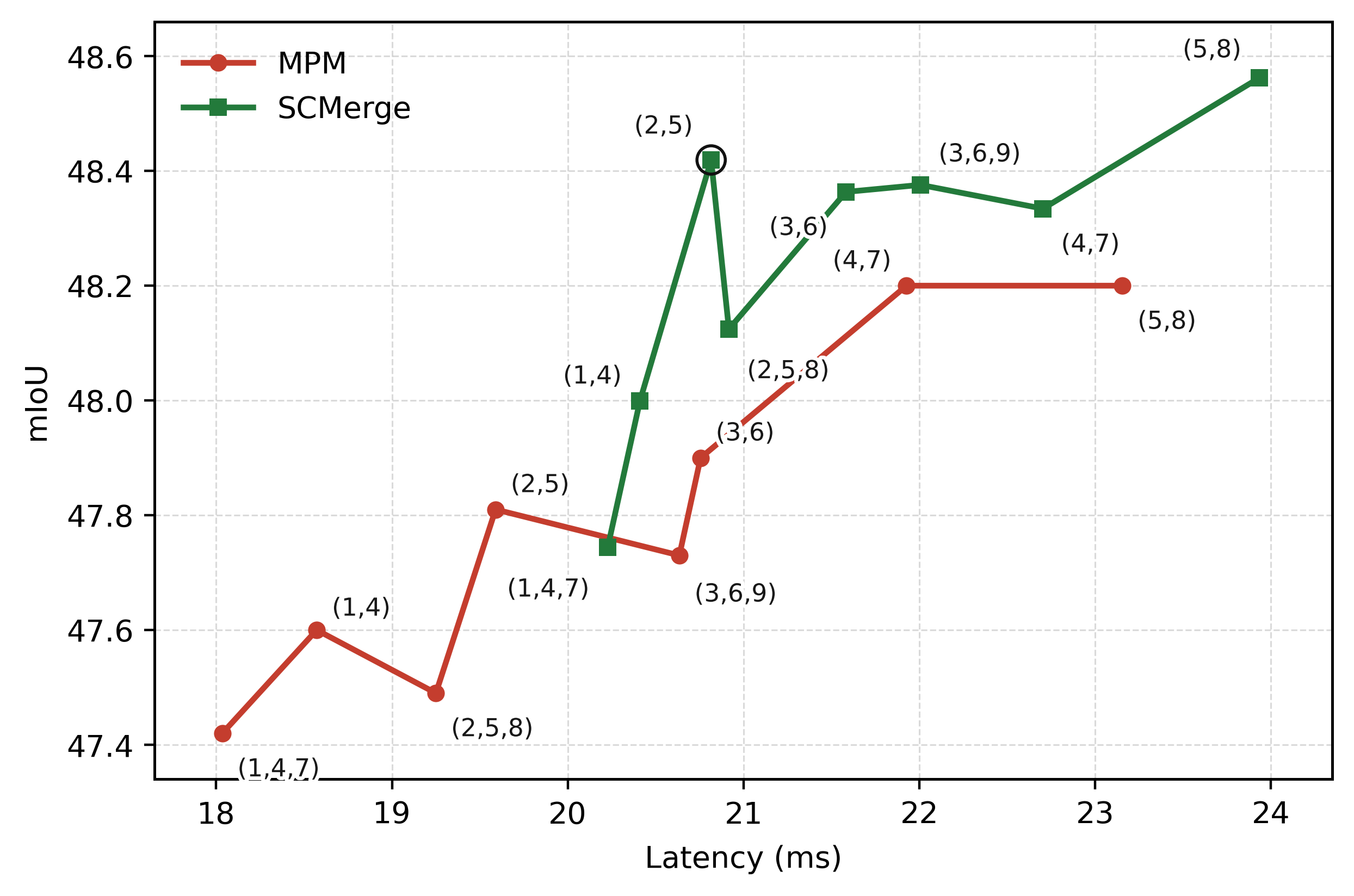}
\caption{Accuracy–latency trade-off of MPM and our \method on ADE20K. Labels denote zero-based encoder-block indices. Our \method consistently achieves a higher mIoU and a better accuracy–latency trade-off than MPM.}
\label{fig:schedule_pareto}
\end{figure}

Figure~\ref{fig:schedule_pareto} analyzes the effect of insertion depth on the accuracy–efficiency trade-off. Applying token merging at earlier encoder stages yields greater computational savings in subsequent layers but makes token matching more susceptible to incorrect pairing. In contrast, merging at later stages preserves finer spatial structures and segmentation accuracy at the cost of reduced latency gains. Therefore, token merging is performed in the early-to-middle encoder blocks.

\subsection{Analyses}
\label{sec:mechanism_analysis}

\textbf{Boundary Preservation.}
Table~\ref{tab:boundary_analysis_main} reports B-mIoU@5 (mIoU within 5 pixels of boundaries) and B-Merge (fraction of boundary tokens involved in merges). Our \method improves B-mIoU@5 while sharply reducing B-Merge, confirming that the boundary-sensitivity term $B_{ij}$ prevents boundary tokens from being merged, improving boundary segmentation quality.

\textbf{Neighborhood Consistency.}
Table~\ref{tab:pair_analysis_main} reports the neighborhood consistency of selected token pairs.  Cross-cls denotes the fraction of pairs connecting tokens from different classes, while N-Agree measures the position-wise label agreement between the $3\times3$ neighborhoods of paired tokens. Our \method produces significantly fewer cross-class pairs and achieves higher N-Agree, indicating that $D_{ij}$ effectively promotes the selection of semantically consistent token pairs.

\begin{table}[!t]
\centering
\captionsetup{skip=4pt}
\caption{Effect of the boundary-sensitivity term $B_{ij}$ on ADE20K.}
\label{tab:boundary_analysis_main}
\footnotesize
\setlength{\tabcolsep}{4.2pt}
\begin{tabular}{lcc}
\toprule
Method & B-mIoU@5 $\uparrow$ & B-Merge (\%) $\downarrow$ \\
\midrule
Full tokens & 35.37 & -- \\
MPM & 34.22 & 50.33 \\
MPM-top & 34.59 & 42.55 \\
\method & \textbf{35.29} & \textbf{22.55} \\
\bottomrule
\end{tabular}
\end{table}

\begin{table}[!t]
\centering
\captionsetup{skip=4pt}
\caption{Effect of the neighborhood consistency term $D_{ij}$ on ADE20K.}
\label{tab:pair_analysis_main}
\footnotesize
\setlength{\tabcolsep}{4.2pt}
\begin{tabular}{lcc}
\toprule
Method & Cross-cls (\%) $\downarrow$ & N-Agree (\%) $\uparrow$ \\
\midrule
MPM & 3.874 & 93.28 \\
MPM-top & 3.252 & 94.01 \\
\method & \textbf{2.168} & \textbf{95.25} \\
\bottomrule
\end{tabular}
\end{table}

%% file: sec/5_conclusion.tex
\section{Conclusion}
\label{sec:conclusion}

In this work, we presented SCMerge, a training-free token merging method that conducts token merging based on both feature similarity and spatial consistency for semantic segmentation. The method is flexible and can be plugged into existing ViT-based semantic segmentation models for improving computation efficiency with less quality drop. Extensive comparison experiments on three benchmark semantic segmentation datasets  ADE20K, Pascal Context, and Cityscapes demonstrate that our SCMerge can achieve better accuracy-latency trade-off compared with existing state-of-the-art training-free token merging methods. Although our SCMerge with the spatial-consistency penalty mitigates boundary blurring, very small objects and extremely thin structures may still be smoothed due to information aggregation. In addition, since SCMerge is training-free and introduces no learnable parameters, its merging decisions are based on fixed, hand-designed criteria. This may limit its adaptability to diverse or unseen visual patterns. Addressing these limitations is a direction for our future work while preserving the plug-and-play and training-free advantages of SCMerge.